\documentclass{article}

\PassOptionsToPackage{numbers, square}{natbib}

\usepackage[preprint]{neurips_2024}

\usepackage{booktabs}
\usepackage{geometry}
\usepackage[utf8]{inputenc} %
\usepackage[T1]{fontenc}    %
\usepackage{hyperref}       %
\usepackage{url}            %
\usepackage{booktabs}       %
\usepackage{amsfonts}       %
\usepackage{nicefrac}       %
\usepackage{microtype}      %
\usepackage{xcolor}         %
\usepackage{natbib}
\usepackage{graphicx}
\usepackage{tikz}
\usetikzlibrary{positioning, shapes.geometric, arrows}
\usetikzlibrary{positioning,calc,bayesnet,external}
\usepackage{pgfplots}
\usepackage{booktabs}
\usepackage{geometry}
\usepackage{adjustbox}

\RequirePackage[noend]{algorithmic}
\usepackage[ruled]{algorithm}
\usepackage{amsmath}

\usepackage{graphicx}
\usepackage{subcaption}
\usepackage{wrapfig}

\newcommand{\conf}{\mathbf{x}}

\definecolor{darkorchid}{rgb}{0.6, 0.2, 0.8}

\title{Improving Hyperparameter Optimization with Checkpointed Model Weights}

\author{%
  Nikhil Mehta, Jonathan Lorraine, Steve Masson, Ramanathan Arunachalam\\
  \textbf{Zaid Pervaiz Bhat, James Lucas, Arun George Zachariah}\\
  NVIDIA\\
  \url{https://research.nvidia.com/labs/toronto-ai/FMS/}\\
}

\begin{document}

\maketitle

\begin{abstract}
    When training deep learning models, the performance depends largely on the selected hyperparameters.
    However, hyperparameter optimization (HPO) is often one of the most expensive parts of model design.
    Classical HPO methods treat this as a black-box optimization problem.
    However, gray-box HPO methods, which incorporate more information about the setup, have emerged as a promising direction for more efficient optimization.
    For example, using intermediate loss evaluations to terminate bad selections.
    In this work, we propose an HPO method for neural networks using logged checkpoints of the trained weights to guide future hyperparameter selections.
    Our method, Forecasting Model Search (FMS), embeds weights into a Gaussian process deep kernel surrogate model, using a permutation-invariant graph metanetwork to be data-efficient with the logged network weights.
    To facilitate reproducibility and further research, we open-source our code.
    \footnote{\url{https://github.com/NVlabs/forecasting-model-search}}
\end{abstract}

\section{Introduction}
\label{Introduction}

Machine learning models have many design choices, or hyperparameters, which significantly affect the model's final performance \cite{bengio2012practical, hutter2019automated}.
These hyperparameters include optimization parameters (e.g., learning rate), architectural parameters (e.g., model selection), regularizers, data augmentation strategies, and many more~\citep{yu2020hyper}.
The hyperparameter selection often governs model quality, training speed, and generalization to unseen data.
Hyperparameter optimization (HPO) is crucial for achieving high-quality results with deep learning models.
However, optimizing hyperparameters is challenging due to the infeasibility of gradient-based optimization and the large, complex search space \cite{bergstra2012random, bischl2019hyperparameter}.
Existing HPO methods are often computationally expensive and time-consuming, making them impractical for many real-world applications where resources and time are limited \cite{snoek2012practical, thornton2013auto}.
Efficient HPO is essential for pushing the boundaries of what machine learning models can achieve.

Classical HPO methods treat the model's final performance evaluation as a black-box function, ignoring the underlying optimization process.
Grid and random search methods do not leverage any information about the training process, leading to less efficient searches through the hyperparameter space \cite{bergstra2012random, hastie2009elements}.
More sophisticated black-box techniques, like standard Bayesian Optimization (BO), model the objective function probabilistically to better decide which hyperparameters to evaluate next.
While more efficient, these methods ignore valuable information generated during the training process \cite{snoek2012practical}.
Multifidelity methods have emerged as a promising direction for improving HPO efficiency.
They go beyond black-box optimization by using lower cost and fidelity objective evaluations, like performance early in training, to inform higher cost and fidelity evaluations.
For example, Hyperband \cite{li2018hyperband} employs a bandit-based approach to dynamically allocate resources based on intermediary loss evaluations, significantly speeding optimization.
HPO methods are enhanced by making assumptions about the problem structure beyond merely being a black box.
\newpage

First, we are interested in HPO methods that work well for the common and important design choice: choosing which powerful pretrained model to start with, for example, from a model hub \cite{wolf2020transformers}.
Current HPO techniques struggle with this paradigm because they treat the model selection as an extra categorical hyperparameter, overlooking critical information about the models, such as their architecture and weights~\cite{arango2024quicktune}.
Techniques like LogME~\cite{pmlr-v139-you21b} and LEEP~\cite{nguyen2020leep} attempt to identify the best pretrained model from a hub but still require hyperparameter optimization after the model has been chosen.
This sequential process is expensive as each model’s hyperparameters must be tuned individually.
QuickTune \cite{arango2024quicktune} addresses this by jointly optimizing model selection with other hyperparameters but is limited to a categorical model embedding.

Second, we desire so-called foundational HPO methods, which can learn from a large corpus of data from hyperparameter evaluations in various settings, including varied architectures, datasets, losses, and hyperparameter search spaces.
OptFormer \cite{chen2022learning} was one of the first successful foundational HPO methods, training on many loss trajectories with varied hyperparameters to improve performance.
DyHPO also shows promise as a foundational HPO method by allowing training on a large corpus of existing evaluations and integrating learning curves into a multifidelity GP-based Bayesian optimization framework.
Despite these advancements, both methods still leave valuable information on the table, such as logged checkpoints of neural networks during training, which is a rich source of information about the architecture, loss, training data, and optimization process.

This work proposes an HPO method that (a) works well for model selection from hubs (see Section~\ref{sec:exp_1}) and (b) allows training on a large corpus of existing hyperparameter evaluation metadata, including logged network weights (see Section~\ref{sec:exp_2}).
Our approach, Forecasting Model Search (FMS), builds on DyHPO by embedding logged network weights into a Gaussian process deep kernel surrogate model, using a permutation-invariant graph metanetwork to be more data-efficient with the logged network weights.
This method provides a more effective way to optimize hyperparameters, particularly in scenarios involving pretrained model selection and fine-tuning.

\begin{figure}[t!]
  \centering
  \resizebox{0.99\textwidth}{!}{\begin{tikzpicture}[
  node distance=1.5cm,
  >=stealth',
  every node/.style={align=center, font=\fontsize{9}{0}\selectfont},
]

\tikzstyle{boxstyle}=[rectangle,draw=black,minimum size=15]
\tikzstyle{circlestyle}=[circle,draw=black,minimum size=15]
\tikzstyle{roundedrect}=[rounded rectangle,draw=black,minimum size=15]
\tikzstyle{blueroundedrect}=[rounded rectangle,draw=blue,minimum size=15]
\tikzstyle{bluerectangle}=[rectangle,draw=blue,minimum size=15]
\tikzstyle{bluearrow}=[->, draw=blue]

\node[boxstyle](x_input) at (0, 2) {Hyperparameter Config $\textbf{x}_{i}$};
\node[boxstyle](b_input) at (0, 1) {Budget $j$};
\node[boxstyle](lc_input) at (0, 0) {Learning Curve $\mathbf{Y}_{i,j-1}$};
\node[bluerectangle](pigm_input) at (0, -1) {\textcolor{blue}{Model Checkpoint $\mathbf{W}$}};

\node[circlestyle](layer1) at (3, 1.5) {Hidden\\Layer 1};
\node[circlestyle](layer2) at (7, 1.5) {Hidden\\Layer 2}; 

\node[roundedrect](conv) at (3.5, 0) {Conv}; 
\node[roundedrect](max) at (5, 0) {Max}; 
\node[blueroundedrect](pigm) at (3.5, -1) {\textcolor{blue}{PIGMN}};
\node[blueroundedrect](readout) at (6, -1) {\textcolor{blue}{Readout}};

\draw[->] (x_input) -- (layer1);
\draw[->] (b_input) -- (layer1);
\draw[->] (lc_input) -- (conv);
\draw[bluearrow] (pigm_input) -- (pigm);
\draw[->] (layer1) -- (layer2);
\draw[->] (conv) -- (max);
\draw[->] (max) -- (layer2);
\draw[bluearrow] (pigm) -- (readout);
\draw[bluearrow] (readout) -- (layer2);

\node[boxstyle, right=of layer2] (output) {Features for\\Deep Kernel GP $\psi$};
\draw[->] (layer2) -- (output);

\node[boxstyle, below=of output, yshift=1cm] (kernelOutput) {Kernel $\mathbf{K}$ from Equation~\ref{eq:weight_kernel}};
\draw[->] (output) -- (kernelOutput);

\node[boxstyle, below=of kernelOutput, yshift=1cm] (mean_var_Output) {Predictive mean and variance\\ $\mu(\mathbf{x}_{i}), \sigma^2(\mathbf{x}_{i})$ from Equation~\ref{eq:pred_mean_var}};
\draw[->] (kernelOutput) -- (mean_var_Output);

\end{tikzpicture}}
  \caption{
      We show an overview of our method, Forecasting Model Search (FMS), which builds on DyHPO's multifidelity method from Algorithm~\ref{alg:dyhpo}.
      Novel components of FMS are highlighted in {\color{blue}blue} and further detailed in Algorithm~\ref{alg:fms}.
      We include DyHPO's features from the hyperparameter configuration, budget, and learning curve \cite{wistuba2023supervising}.
      Notably, we also featurize the model's checkpointed weights $\mathbf{W}$ with a permutation-invariant graph metanetwork (PIGMN) as in Section~\ref{pigms} for input to a deep kernel GP (see Equation~\ref{deepkernelgp}/\ref{eq:weight_kernel}).
      This provides the HPO with an -- often pre-existing -- rich source of information, which implicitly includes the architecture, dataset, loss, and optimization process.
      FMS shows improved predictions about hyperparameter performance across compute budgets (see Table~\ref{ktau-values}), improved quality of the final selected configuration across compute budgets (see Figure~\ref{fig:regret_over_time}), and a potential to generalize beyond what was seen in training (see Figure~\ref{fig:fms-transfer}).
      Specific design choices for this surrogate model are detailed in Appendix Section~\ref{meta-hyperparams}.
  }
  \label{fig:fms-overview-tikz}
\end{figure}
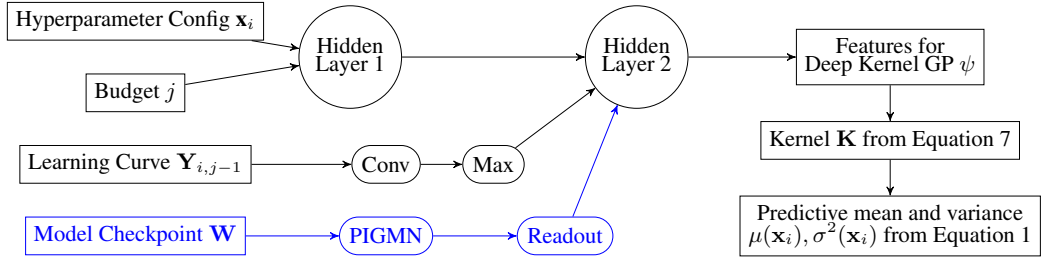

Our contributions include:
\begin{enumerate}
    \item Introducing Forecasting Model Search (FMS), a novel and effective HPO method building on DyHPO by also leveraging logged model weights, outlined in Figure~\ref{fig:fms-overview-tikz}.
    \item Empirically improving performance on varied benchmarks, as in Table~\ref{ktau-values}, and Figures~\ref{fig:regret_over_time} \& \ref{fig:fms-transfer}.
    \item Providing open-source code, allowing others to reproduce our experiments \& build on our method easily.
\end{enumerate}

\newpage
\section{Background}\label{sec:background}
We include a summary of our notation in Appendix Table~\ref{tab:TableOfNotation}.
This section starts by describing the Bayesian Optimization (BO) HPO framework.
Next, we briefly cover the Gaussian processes (GPs) we use for our surrogate model during BO.
Finally, we cover DyHPO, a multifidelity BO variant for HPO.
After, in Section~\ref{Method}, we describe our method FMS, which builds on DyHPO by also conditioning on checkpoints of logged weights, which are featurized with a graph meta-network (GMN).

\textbf{Bayesian Optimization (BO)} is an approach for optimizing noisy, expensive, and non-differentiable functions, particularly useful in HPO.
It employs a probabilistic model, commonly a Gaussian Process (GP), as a surrogate to approximate the expensive objective function \( f: \mathcal{X} \to \mathbb{R} \).

The core idea is to use a surrogate model to make informed decisions about where the function should be evaluated next in the hyperparameter space, aiming to improve model performance with minimal computational expense.
The next query point \( \mathbf{x}_{*} \) is selected by (approximately) maximizing\footnote{The $\arg\max$ could have non-unique solutions, but we use notation as if unique for simplicity.} an acquisition function \( a \), which is designed to balance the exploration of less known regions with the exploitation of promising areas:
\[
\mathbf{x}_{*} \approx \arg\max_{\mathbf{x} \in \mathcal{X}} a(\mathbf{x} | \mathcal{D}),
\]
where \(\mathcal{D} = \{(\mathbf{x}_i, y_i)\}_{i=1}^n\) consists of previously evaluated hyperparameter configurations \(\mathbf{x}_i\) and their corresponding observed performances \(y_i = f(\mathbf{x}_i)\).
Our acquisition function is shown in Equation~\ref{eq:ei_acquistion}.

\textbf{Gaussian Processes (GPs)} offer a robust way to model the relationship between hyperparameter configurations and a model's performance.
A GP assumes that the performance metrics in the hyperparameter space have a joint Gaussian distribution, characterized by:
\[
f(\mathbf{x}) \sim \mathcal{GP}(m(\mathbf{x}), k(\mathbf{x}, \mathbf{x}')),
\]
where \( m(\mathbf{x}) \) is the mean function, typically set to zero, and \( k(\mathbf{x}, \mathbf{x}') \) is the covariance function, encapsulating assumptions about the function’s smoothness and variability.
After observing data \(\mathcal{D}\), the GP provides a posterior function distribution, quantifying uncertainty and guiding the selection of the next point to evaluate.
The predictive mean and variance at a new point \(\mathbf{x}_{*}\) are given by:
\begin{equation}\label{eq:pred_mean_var}
\mu(\mathbf{x}_{*}) = \mathbf{k}_{*}^\top (\mathbf{K} + \sigma^2_n \mathbf{I})^{-1} \mathbf{y},
\hspace{0.07\textwidth}
\sigma^2(\mathbf{x}_{*}) = k(\mathbf{x}_{*}, \mathbf{x}_{*}) - \mathbf{k}_{*}^\top (\mathbf{K} + \sigma^2_n \mathbf{I})^{-1} \mathbf{k}_{*},
\end{equation}
where \(\mathbf{k}_{*}\) is the vector of covariances between \(\mathbf{x}_{*}\) and the training inputs, \(\mathbf{K}\) is the training input covariance matrix, \(\mathbf{y}\) is the observed performance vector, and \(\sigma^2\) is the noise variance.
This probabilistic framework benefits HPO, allowing sequential decision-making under uncertainty and targeting regions in the hyperparameter space predicted to yield the largest improvements.

\subsection{Dynamic Multifidelity Hyperparameter Optimization (DyHPO)}
DyHPO~\citep{wistuba2023supervising} leverages GPs with deep kernels in a BO framework to optimize hyperparameters efficiently using evaluations at varying fidelity levels.
Lower-fidelity, less computationally expensive evaluations provide preliminary insights that inform more costly, high-fidelity evaluations.
This approach allows for efficient exploration and exploitation across different computational budgets, making the optimization process more resource-efficient and scalable.

GPs with deep kernels are used as DyHPO's surrogate model, capturing complex relationships between hyperparameters, budget levels, and performance metrics.
The kernel is defined as:
\begin{equation}
\label{deepkernelgp}
    \mathbf{K}(\boldsymbol{\theta}, \mathbf{w}, \mathcal{D}) := k(\varphi(\boldsymbol{\conf}_i, \mathbf{Y}_{i,j-1}, j; \mathbf{w}), \varphi(\boldsymbol{\conf}_{i'}, \mathbf{Y}_{i',j'-1}, j'; \mathbf{w}); \boldsymbol{\theta}),
\end{equation}
where $\varphi$ represents a neural network transforming hyperparameters $\boldsymbol{\conf}_i$ and learning curves $\mathbf{Y}_{i,j-1}$ into a feature space (as in Figure~\ref{fig:fms-overview-tikz}), and $k$ is a base kernel function such as the squared exponential kernel.
The notation $\mathbf{Y}_{i,j-1}$ represents the logged performance metrics observed up to budget level $j-1$ for the $i^{\textnormal{th}}$ configuration.
Here, $\mathcal{D}$ represents tuples of (hyperparameter $\boldsymbol{\conf}_i$, budget $j$, loss trajectory $\mathbf{Y}_{i,j-1}$).

The parameters of the kernel, $\boldsymbol{\theta}$ (ex., the length-scale \(\ell\)), and the neural network weights $\mathbf{w}$, are learned jointly by maximizing the data likelihood using gradient-based optimizers like Adam \cite{kingma2014adam}.
The loss function is the GP's negative log marginal likelihood (NLML), combining a data fit and complexity penalty term,  is given by:
\begin{equation}
\mathcal{L}(\mathcal{D}) = \frac{1}{2} \mathbf{y}^\top \mathbf{K}(\boldsymbol{\theta}, \mathbf{w}, \mathcal{D})^{-1} \mathbf{y} + \frac{1}{2} \log \left|\mathbf{K}(\boldsymbol{\theta}, \mathbf{w}, \mathcal{D})\right| + \frac{n}{2} \log 2\pi,
\end{equation}

where $\mathbf{y}$ is the vector of observed performance metrics, $\mathbf{K}(\boldsymbol{\theta}, \mathbf{w}, \mathcal{D})$ is the kernel matrix, and $n$ is the number of observations.
The gradient of the NLML with respect to the parameters $\boldsymbol{\theta}$ and $\mathbf{w}$ is:
\begin{equation}\label{eq:kernel_grad}
\nabla_{\boldsymbol{\theta}, \mathbf{w}} \mathcal{L}(\mathcal{D}) = -\left(\mathbf{y}^\top \mathbf{K}(\boldsymbol{\theta}, \mathbf{w}, \mathcal{D})^{-1} \mathbf{y} - \text{Tr}\left(\mathbf{K}(\boldsymbol{\theta}, \mathbf{w}, \mathcal{D})^{-1}\right)\right)
\end{equation}
While training the surrogate model in DyHPO, the entire dataset $\mathcal{D}$ is used for each evaluation of the GP surrogate, limiting scalability to larger datasets.
Using mini-batches during training or inference could improve scalability but pose difficulties for accurately evaluating vanilla GPs.
When scaling to big datasets, using ultra-scalable GPs~\citep{wang2022scalable} or other more scalable surrogates~\citep{antoran2024scalable} may be required.

\paragraph{Algorithm Overview.}
DyHPO iteratively selects hyperparameter configurations and corresponding budgets to evaluate by maximizing a multifidelity version of the Expected Improvement (EI) acquisition function.
The budget is defined as the number of epochs used for evaluation, simplifying the process.
This approach allows DyHPO to dynamically allocate resources at varying budget levels based on insights from previous evaluations, focusing computational efforts on the most promising configurations.
The multifidelity EI is defined as in \citet{swersky2013multi} and \citet{wistuba2023supervising}:
\begin{equation}\label{eq:ei_acquistion}
    \operatorname{EI_{MF}}(\boldsymbol{\conf}, j | \mathcal{D}) = \mathbb{E}\left[\max\left\{f(\boldsymbol{\conf}, j) - y_j^{\text{max}}, 0\right\}\right],
\end{equation}
where $y_j^{\text{max}}$ is the highest observed performance at any given budget $j$.
Here, for simplicity, the budget $j$ is treated as an integer representing the number of epochs and maximized alongside other hyperparameters.
DyHPO's acquisition function optimization, which we share, is described in Appendix Section~\ref{Acquisition Function Maximization}.

\begin{algorithm}[H]
    \caption{DyHPO's Algorithm \cite{wistuba2023supervising}}
    \label{alg:dyhpo}
    \begin{algorithmic}[1]
        \STATE Initialize $\mathcal{D}$ with any preexisting evaluations of configurations $\conf_{\cdot}$, losses $\mathbf{Y}_{\cdot,j-1}$, and budgets $j$ and update the GP model's parameters $\boldsymbol{\theta}$ and $\mathbf{w}$ using gradient-based optimization with $\nabla_{\boldsymbol{\theta}, \mathbf{w}} \mathcal{L}$ from Equation~\ref{eq:kernel_grad}
        \WHILE{computational budget not exhausted}
        \STATE Select configuration and budget $(\boldsymbol{\conf}_{i}, j)$ by maximizing $\operatorname{EI_{MF}}(\boldsymbol{\conf}, j | \mathcal{D})$
        \STATE Evaluate configuration with budget: $y_{i} = f(\boldsymbol{\conf}_{i}, j)$, with loss trajectory $\mathbf{Y}_{i,j-1}$
        \STATE Update $\mathcal{D}$ with the new observation $(\boldsymbol{\conf}_{i}, j, \mathbf{Y}_{i,j-1})$
        \STATE Update the GP model's parameters $\boldsymbol{\theta}$ and $\mathbf{w}$ using gradient-based optimization for $N$ steps or until termination criteria are satisfied with $\nabla_{\boldsymbol{\theta}, \mathbf{w}} \mathcal{L}$ from Equation~\ref{eq:kernel_grad}
        \ENDWHILE
        \STATE \textbf{return} configuration $\boldsymbol{\conf}_{i}$ with the best observed performance $y_{i}$
    \end{algorithmic}
\end{algorithm}

DyHPO's method in Algorithm~\ref{alg:dyhpo} dynamically adjusts to observed performance data, allocating more resources to promising configurations as more data is gathered, thereby optimizing the use of computational resources across different budget levels.
Training DyHPO on a large preexisting set of evaluations can enhance its performance through transfer learning, enabling better generalization.
Alternatively, DyHPO can dynamically generate its dataset, continuously improving as more data is gathered during optimization.

\newpage
\section{Our Method: Forecasting Model Search (FMS)}\label{Method}
\vspace{-0.01\textheight}

DyHPO's method trains a network to featurize an optimization problem to guide HPO by inputting evaluated hyperparameter and loss-trajectories and outputting features for a GP surrogate by training on a large (and progressively growing) dataset of hyperparameter evaluations.
However, this provides limited context for our optimization problem.
Our method provides additional context for the GP to condition on by featurizing the neural network weights.
These weights are stored at intermediary checkpoints during optimization and encode information about the architecture, loss, training dataset, and optimization process.
We describe the permutation-invariant graph metanetwork (PIGMN) in Section~\ref{pigms}, a special graph neural network customized to be data efficient on neural network inputs, which we use to featurize the weights.
Then, we describe the entire augmented architecture and HPO procedure with checkpointing in Section~\ref{sec:method_fms}.

\vspace{-0.01\textheight}
\subsection{Permutation-Invariant Graph Metanetworks (PIGMNs)}\label{pigms}
\vspace{-0.01\textheight}

In the DyHPO framework, we use permutation-invariant graph metanetworks (PIGMNs) to manage the diverse architectures during optimization.
PIGMNs ensure that outputs are invariant to the input graph nodes' permutations, using network symmetries to enhance training data efficiency~\citep{lim2023graph}.
Key to the PIGMN is constructing an input graph of the checkpointed neural network weights $\mathbf{W}$, denoted $\mathcal{G}^{(0)}(\mathbf{W})$, where each node corresponds to a layer, and edges represent the weights connecting these layers.
Further details are in \citet{lim2023graph}.
PIGMNs use convolutional graph layers with permutation-invariant operations to process $\mathcal{G}^{(0)}$  to generate features $\xi$:
\vspace{-0.005\textheight}
\begin{align}
    \xi(\mathcal{G}^{(0)}) = \sum_{\substack{v \in V\left(\mathcal{G}^{L}\right)}} \frac{ \mathbf{h}_v^{L}}{| V\left(\mathcal{G}^{L}\right) |} \textnormal{, where } \mathcal{G}^{(l+1)} = \sigma \left( \sum_{k=1}^{K} \Theta_k^{(l)} \ast \mathcal{G}^{(l)} \right) \textnormal{ for } l = 0, \dots, L-1
\end{align}
where $\ast$ denotes the graph convolution operation, $\Theta_k^{(l)}$ represents the trainable parameters at layer $l$ for the $k^{\textnormal{th}}$ kernel, and $\sigma$ is an activation function.
$V(\mathcal{G})$ is the set of nodes in the graph $\mathcal{G}$ and $\mathbf{h}_v^{L}$ is the feature vector of node $v \in V$ in the final layer $L$.
This ensures that the feature vector captures the information from all graph nodes and is invariant to permutations.
Notably, GMNs can input weights from models with varying architectures, making our method more generalizable.

\vspace{-0.01\textheight}
\subsection{Combining DyHPO with PIGMNs for FMS}\label{sec:method_fms}
\vspace{-0.01\textheight}
FMS enhances the DyHPO framework by using PIGMNs to featurize model weights, encoding information about the architecture and the dataset the model was trained on.
Our method is designed to be effective when selecting and fine-tuning models from a model hub because the input weights incorporate information about the architecture that existing HPO methods ignore.
We add a PIGMN to DyHPO's architecture to encode the model weights $\mathbf{W}$ into an augmented kernel function:
\vspace{-0.0025\textheight}
\begin{equation}\label{eq:weight_kernel}
    \mathbf{K}(\boldsymbol{\theta}, \mathbf{w}) := k(\psi(\boldsymbol{\conf}_i, {\color{blue}\mathbf{W}_i}, \mathbf{Y}_{i,j-1}, j; \mathbf{w}), \psi(\boldsymbol{\conf}_{i'}, {\color{blue}\mathbf{W}_{i'}}, \mathbf{Y}_{i'\!,j'-1}, j'; \mathbf{w}); \boldsymbol{\theta}),
\end{equation}
where $\psi$ is a feature extractor for hyperparameters $\boldsymbol{\conf}$, learning curves $\mathbf{Y}_{i,j-1}$, budgets $j$, and -- the key difference from Equation~\ref{deepkernelgp} -- model weights $\mathbf{W}$ in {\color{blue}blue}.
We use the same multifidelity EI acquisition function as DyHPO from Equation~\ref{eq:ei_acquistion}.
Algorithm~\ref{alg:fms} shows our entire method, highlighting differences from DyHPO.
We provide the saved weights $\textbf{W}$ as input to the featurizer, pretrain our featurizers with any available logged weight checkpoints, and ensure we save the weights from any prescribed hyperparameter evaluations.
Many HPO pipelines already checkpoint the model weights to avoid retraining once the final, optimized hyperparameters $\conf$ are found.

\begin{algorithm}[H]
    \caption{Our Forecasting Model Search (FMS) method, with changes from Algorithm~\ref{alg:dyhpo} in \textcolor{blue}{blue}.}
    \label{alg:fms}
    \begin{algorithmic}[1]
        \STATE Initialize $\mathcal{D}$ with any preexisting evaluations of configurations $\conf$, , losses $\mathbf{Y}_{i,j-1}$, \textcolor{blue}{their weights $\mathbf{W}$,} and budgets $j$ and update the GP parameters $\boldsymbol{\theta}$ and $\mathbf{w}$ via gradient-based optimization with $\nabla_{\boldsymbol{\theta}, \mathbf{w}} \mathcal{L}$ from Equation~\ref{eq:kernel_grad}
        \WHILE{computational budget not exhausted}
        \STATE Select $(\conf_i, \textcolor{blue}{\mathbf{W}_i}, j)$ by maximizing $\operatorname{EI_{MF}}(\conf, \textcolor{blue}{\mathbf{W}}, j | \mathcal{D})$
        \STATE Evaluate configuration and budget: $y_i = f(\conf, \textcolor{blue}{\mathbf{W}}, j)$ , with loss trajectory $\mathbf{Y}_{i,j-1}$
        \STATE Update $\mathcal{D}$ with $(\conf_i, \textcolor{blue}{\mathbf{W}_i}, j, \mathbf{Y}_{i,j-1})${\color{blue}, effectively checkpointing the weights $\mathbf{W}$}
        \STATE Optimize GP parameters $\boldsymbol{\theta}$ and $\mathbf{w}$ via gradient-based optimization for $N$ steps or until termination criteria are satisfied with $\nabla_{\boldsymbol{\theta}, \mathbf{w}} \mathcal{L}$ from Equation~\ref{eq:kernel_grad}
        \ENDWHILE
        \STATE \textbf{return} configuration $\boldsymbol{\conf}_{i}$ with the best observed performance $y_{i}${\color{blue}, and its learned parameters $\mathbf{W}_i$}
    \end{algorithmic}
\end{algorithm}\vspace{-.025\textheight}

\newpage
\section{Experiments}
\label{Experiments}

Our experiments aim to evaluate FMS's compute budget versus quality trade-off and generalization to unseen datasets and architectures.
First, Section~\ref{sec:exp_1} examines the trade-off between HPO performance and total allocated compute budget.
Then, Section~\ref{sec:exp_2} examines FMS's generalization to different datasets and architectures, focusing on the setup of selecting models for fine-tuning.

We use two model hubs and create corresponding benchmarks for fine-tuning models with different hyperparameter configurations.
See Appendix Section~\ref{Generating the Pretrained Model Hub} for the detailed procedure.
The first model hub of \citet{unterthiner2020predicting}, referred to as \emph{Simple CNN Hub}, contains a single CNN architecture with different initializations.
The second hub, called the \emph{Pretrained Model Hub}, is ours and consists of various architectures, including ResNet~\citep{resnet}, ViT~\citep{vit}, CNN~\citep{726791}, and Deep Set~\citep{zaheer2017deep}.
These architectures are pretrained on ImageNet~\cite{imagenet} for classification and later fine-tuned on CIFAR-10 and SVHN (Section~\ref{sec:exp_1}).

Our experiments are designed for the paradigm where users select which pretrained model to fine-tune.
For the \emph{Simple CNN Hub}, pretrained models vary by weight initialization and architectural choices such as the number of layers or hidden units.
With \emph{Pretrained Model Hub}, the architectures to choose from include various ResNets, ViTs, CNNs, and Deep Set variants.
Note that each architecture has multiple pretrained models of different sizes and initializations, which our HPO chooses between.
In general, if we have $N$ models from a hub, we can encode the model choice as a one-hot hyperparameter while -- crucially -- the checkpointed weights are given to the PIGMN, allowing the encoding of relevant information about the architecture, dataset, or loss for the surrogate.

Further details are provided for the experimental procedure in Appendix Section~\ref{Experimental Procedure}, and generating the hubs, datasets, corresponding benchmarks, and hyperparameter search space in Appendix Section~\ref{Hyperparameter Settings}.
The design choices for the surrogate model itself can be found in Appendix Section~\ref{meta-hyperparams}.

\subsection{FMS for Fine-tuning on a Dataset}\label{sec:exp_1}

We first use Kendall's $\tau$ correlation coefficient to assess the agreement between the rankings from our hyperparameter optimization methods and the actual performance rankings of the configurations.
A higher Kendall's $\tau$ value indicates a better ranking method, with more details in \cite{kendall1938new}.
Table \ref{ktau-values}'s results show Kendall's $\tau$ values at various budgets for different model hubs.
Our results demonstrate that FMS-GMN consistently achieves the highest Kendall's $\tau$ values, indicating superior performance in ranking hyperparameter configurations compared to other methods.
Even the simpler FMS variants outperform traditional methods like DyHPO, Random Search, and GP, suggesting FMS is more effective in predicting the best hyperparameter configurations.
So, in the next experiments, we investigate if this leads to improved performance from the best model found by the HPO.

\begin{table}[ht!]
  \caption{
    Kendall's $\tau$ values at various budgets for different model hubs.
    NFN variants can not process the weights of diverse architectures~\cite{zhou2023permutation}, so they are not run on either PTM hub.
  }
  \label{ktau-values}
  \centering
  \small
  \begin{tabular}{lcc|cc|cc}
    \toprule
    \textbf{Method} & \multicolumn{2}{c}{\textbf{Simple CNN Hub}} & \multicolumn{2}{c}{\textbf{PTM Hub SVHN}} & \multicolumn{2}{c}{\textbf{PTM Hub CIFAR-10}} \\
    \cmidrule(r){2-3} \cmidrule(r){4-5} \cmidrule(r){6-7}
    & \textbf{50 epochs} & \textbf{100 epochs} & \textbf{50 epochs} & \textbf{100 epochs} & \textbf{50 epochs} & \textbf{100 epochs} \\
    \midrule
    FMS-GMN & \textbf{0.88} & \textbf{0.92} & \textbf{0.86} & \textbf{0.90} & \textbf{0.87} & \textbf{0.91} \\
    FMS-NFN & 0.84 & 0.88 & - & - & - & - \\
    FMS-NFN (no CNN) & 0.82 & 0.85 & - & - & - & - \\
    FMS-GMN (no CNN) & 0.80 & 0.84 & 0.79 & 0.83 & 0.80 & 0.85 \\
    FMS-FLAT & 0.78 & 0.82 & 0.77 & 0.81 & 0.78 & 0.83 \\
    FMS-FLAT (no CNN) & 0.76 & 0.81 & 0.75 & 0.80 & 0.77 & 0.82 \\
    DyHPO (no CNN) & 0.60 & 0.63 & 0.61 & 0.64 & 0.62 & 0.65 \\
    DyHPO & 0.74 & 0.77 & 0.73 & 0.76 & 0.74 & 0.78 \\
    GP & 0.70 & 0.73 & 0.69 & 0.72 & 0.70 & 0.74 \\
    \bottomrule
  \end{tabular}
\end{table}

\begin{figure}[ht!]
  \centering
  \begin{tikzpicture}
    \node (img) {\includegraphics[width=0.97\linewidth]{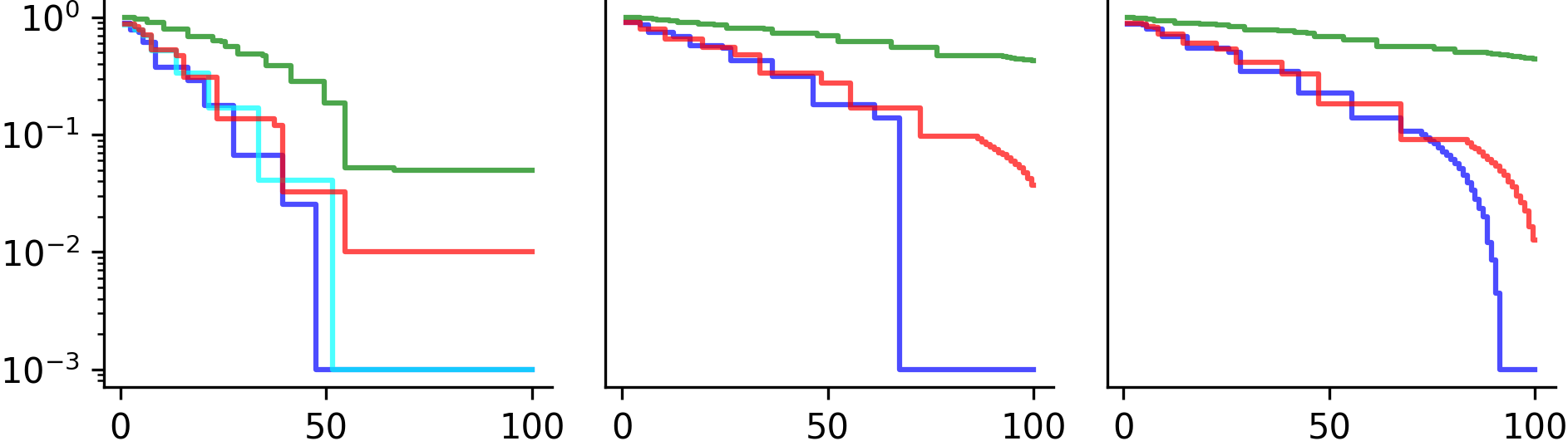}};
    \node[left=of img, node distance=0cm, rotate=90, xshift=0.65cm, yshift=-.9cm, font=\color{black}] {Regret};
    \node[below=of img, node distance=0cm, xshift=-.1cm, yshift=1.1cm,font=\color{black}] {Compute Budget in Epochs};
    \node[above=of img, node distance=0cm, xshift=-4.0cm, yshift=-1.0cm,font=\color{black}] {Simple CNN Hub};
    \node[above=of img, node distance=0cm, xshift=0.5cm, yshift=-1.0cm,font=\color{black}] {PTM Hub (SVHN)};
    \node[above=of img, node distance=0cm, xshift=5.0cm, yshift=-1.0cm,font=\color{black}] {PTM Hub (CIFAR-10)};
    
    \node[below=of img, node distance=2cm, yshift=0.4cm, xshift=0cm] {
      \begin{tikzpicture}
        \node (legend) [draw, fill=white, font=\small, inner sep=3pt] {
          \begin{tabular}{cccc}
            \textcolor{blue}{\rule{0.4cm}{0.2cm}} & FMS-GMN & \textcolor{cyan}{\rule{0.4cm}{0.2cm}} & FMS-NFN \\
            \textcolor{green}{\rule{0.4cm}{0.2cm}} & Random Search & \textcolor{red}{\rule{0.4cm}{0.2cm}} & DyHPO \\
          \end{tabular}
        };
      \end{tikzpicture}
    };
  \end{tikzpicture}
  \caption{
    In each plot, we show the regret against the compute budget across different hubs and various hyperparameter optimization (HPO) methods in each color.
    The regret values reflect the difference between the actual performance and the best possible performance over time.
    Lower regret indicates better performance.
    Our method, FMS-GMN in {\color{blue}blue}, consistently shows lower regret than the strongest baseline DyHPO in {\color{red}red}.
    This persists over most compute budgets across all hubs, demonstrating that our method is effective for HPO.
    FMS-NFN in {\color{cyan}cyan} doesn't support diverse architectures, so it only runs on the \emph{Simple CNN Hub}.
    Figure~\ref{fig:fms-transfer} further investigates the generalization of our FMS-GMN method, while Appendix Figure~\ref{fig:regret_over_time_detailed} shows ablations over our design choices.
  }
  \label{fig:regret_over_time}
\end{figure}
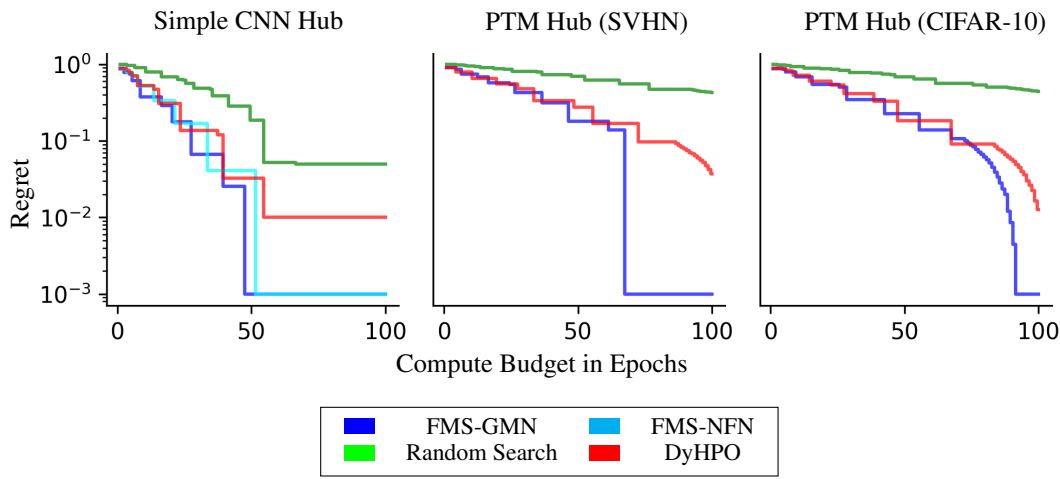

Figure~\ref{fig:regret_over_time} investigates the effectiveness of FMS by recording regret over time in various settings.
Lower regret values indicate better performance with Kendall's $\tau$ coefficient recorded at the $50\textsuperscript{th}$ and $100\textsuperscript{th}$ epochs in Table~\ref{ktau-values}.
Our results show that FMS-GMN achieves the best performance, with consistently lower regret per compute and higher Kendall's $\tau$ values than other methods.

\subsection{Generalization Performance of the Surrogate Model}\label{sec:exp_2}

\begin{figure}[bt!]
  \centering
  \begin{tikzpicture}
    \node (img) {\includegraphics[width=0.97\linewidth]{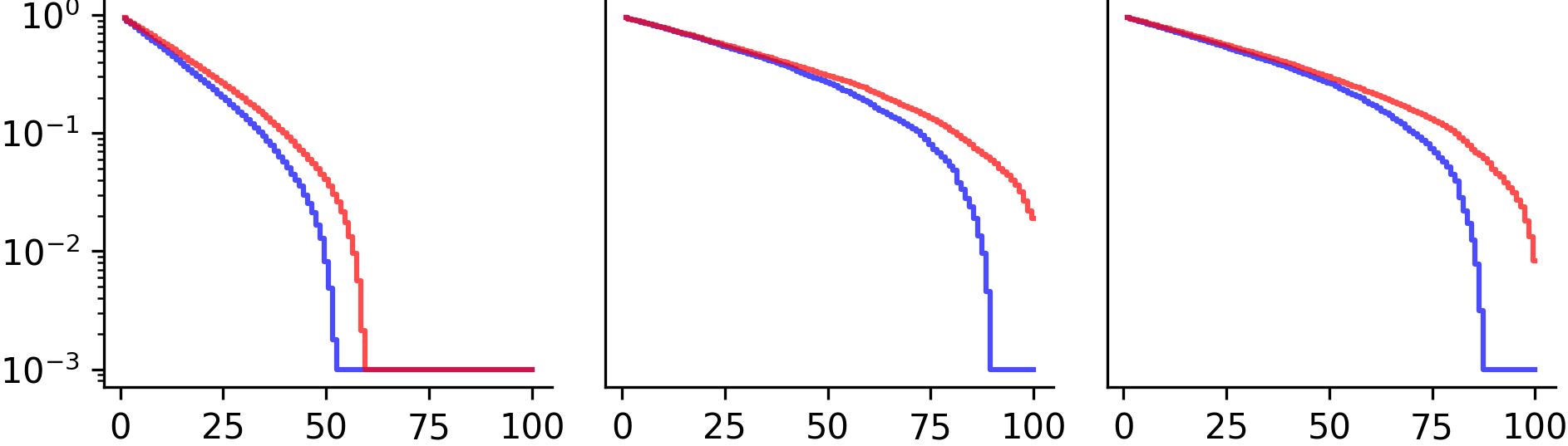}};
    \node[left=of img, node distance=0cm, rotate=90, xshift=0.65cm, yshift=-.9cm, font=\color{black}] {Regret};
    \node[below=of img, node distance=0cm, xshift=-.1cm, yshift=1.1cm,font=\color{black}] {Compute Budget in Epochs};
    \node[above=of img, node distance=0cm, xshift=-4.0cm, yshift=-1.0cm,font=\color{black}] {Simple CNN Hub};
    \node[above=of img, node distance=0cm, xshift=0.5cm, yshift=-1.0cm,font=\color{black}] {PTM Hub (SVHN)};
    \node[above=of img, node distance=0cm, xshift=5.0cm, yshift=-1.0cm,font=\color{black}] {PTM Hub (CIFAR-10)};
    
    \node[below=of img, node distance=2cm, yshift=0.4cm, xshift=0cm] {
      \begin{tikzpicture}
        \node (legend) [draw, fill=white, font=\small, inner sep=3pt] {
          \begin{tabular}{cc}
            \textcolor{blue}{\rule{0.4cm}{0.2cm}} & FMS-GMN generalizing to new datasets \\
            \textcolor{red}{\rule{0.4cm}{0.2cm}} & FMS-GMN trained only on the current dataset \\
          \end{tabular}
        };
      \end{tikzpicture}
    };
  \end{tikzpicture}
  \caption{
    We evaluate the ability of our method to generalize to new datasets and architectures.
    FMS-GMN with generalization shown in {\color{blue}blue} means the model was trained on multiple datasets.
    FMS-GMN without generalization shown in {\color{red}red} was only trained on the current dataset.
    The results show that our model can effectively generalize knowledge between different tasks because the generalization setup's regret is consistently lower than the non-generalization setup, showing it converges faster to a potentially higher-quality solution by leveraging the additional datasets.
  }
  \label{fig:fms-transfer}
\end{figure}
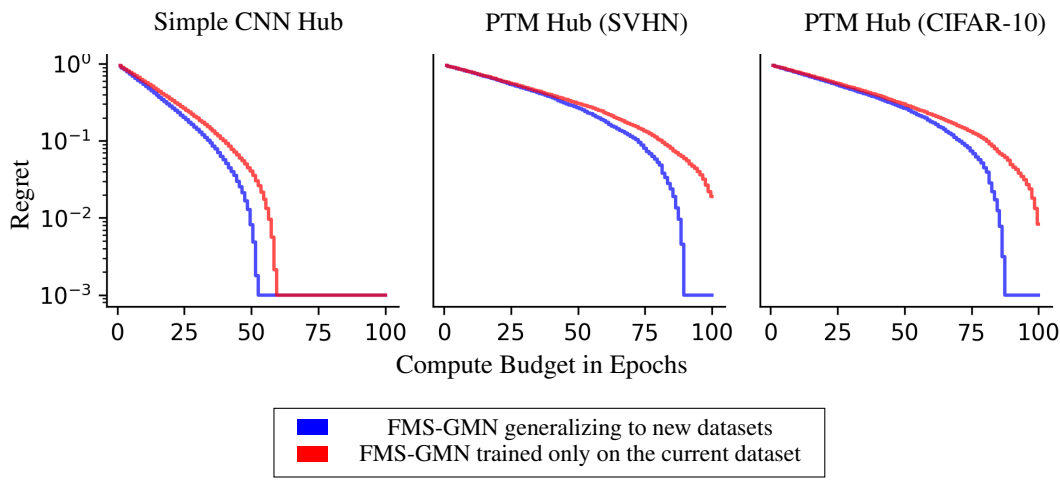

Figure~\ref{fig:fms-transfer} shows the generalization performance of our model on unseen architectures or datasets, measuring performance through regret as before.
FMS-GMN (generalization) is trained on multiple datasets, while FMS-GMN (no generalization) is trained only on one dataset.
Our results show that FMS-GMN can transfer knowledge to improve performance on unseen tasks because the generalization setup's regret is consistently lower than that of the non-generalization setup.

\newpage
\section{Related Work}\label{Related Work}
\vspace{-0.005\textheight}
\subsection{Hyperparameter Optimization (HPO)}
\vspace{-0.005\textheight}

\citet{feurer2019hyperparameter} and \citet{bischl2023hyperparameter} contain a useful introduction to HPO more generally.
Strong HPO methods are critical in AutoML pipelines~\citep{he2021automl, lorraine2022task}, and this is a key use case on which we seek methods that work well.
Initial approaches were largely black-box or model-free, such as grid and random search~\citep{bergstra2012random}.
Subsequent methods include additional problem structure, going beyond black-box optimization to gray-box~\citep{vicol2023bilevel}.
For example, gradient-based HPO, which could be through unrolled-differentiation~\citep{maclaurin2015gradient, raghu2021meta}, implicit differentiation~\citep{franceschi2017forward, lorraine2020optimizing,lorraine2024scalable}, or amortized optimization\citep{lorraine2018stochastic,mackay2018self,bae2020delta,bae2022multi}.
Notably, \citet{raghu2021meta} tune pretraining hyperparameters like us.
Gradient-based methods are extremely scalable but require differentiable objectives, do not apply to our fine-tuning setup or neural architecture search~\citep{elsken2019neural, adam2019understanding, white2023neural}, are often infeasible in use-cases like AutoML.
Our method is a special case of amortized optimization~\citep{amos2022tutorial}, which has been used in varied applications, such as meta-learning~\citep{hospedales2021meta}, 3D generation~\citep{lorraine2023att3d, xie2024latte3d}, optimal transport~\citep{bunne2022supervised} and more.
A novel line of work uses LLMs for HPO by reading the code to implement the model~\citep{zhang2023using}, whereas we directly intake the trained weights.

\textbf{Multifidelity HPO} improves performance by including information about having a finite total compute budget, each hyperparameter evaluation costing a different amount, and low-cost evaluations helping inform the selection of high-cost evaluations.
Hyperband \cite{li2018hyperband} is a multifidelity technique for HPO that selects random hyperparameter configurations with successive halving \cite{jamieson2016non} to focus the budget on more promising configurations while early stopping others.
Methods have been developed to improve Hyperband, such as BOHB~\citep{pmlr-v80-falkner18a}, which constructs a new surrogate for every budget.
The most closely related methods to ours are multifidelity Bayesian optimization HPO, with examples including creating new kernels for multifidelity data \cite{poloczek2017multi, swersky2013multi}, low-fidelity approximations of hyperparameter configurations \cite{swersky2013multi}, and even learning a separate Gaussian process (GP) for each fidelity \cite{kandasamy2016gaussian}.
The later works by \citet{kandasamy2017multi}, \citet{takeno2020multi}, and \citet{wistuba2023supervising} improve on this method by learning one GP for all fidelities.
Importantly, we use the same multifidelity acquisition function setup as DyHPO \cite{wistuba2023supervising}, slowly increasing the invested budget over time.
The core difference between our work and DyHPO is that we also input the neural network weights to improve accuracy in addition to DyHPO's inputs, which include learning curves.

\textbf{Foundational HPO methods} leverage meta-learning and transfer-learning from existing HPO metadata to enhance performance in new settings, using transferable patterns across different datasets, architectures, and tasks.
OptFormer \cite{chen2022learning} uses a transformer-based model trained on diverse optimization trajectories for informed hyperparameter decisions.
Various other approaches include BO for transfer learning~\citep{krause2011contextual, bardenet2013collaborative, poloczek2016warm}, multi-task BO~\citep{wistuba2021few, feurer2018scalable, swersky2013multi, yogatama2014efficient, poloczek2017multi, perrone2018scalable, rothfuss2021pacoh}, learning acquisition functions~\citep{volpp2019meta}, and meta-BO~\citep{feurer2015initializing}
However, these methods often overlook valuable insights from logged training checkpoints, which our method uses.

\textbf{HPO for Model Search}, related to neural architecture search (NAS)~\citep{white2023neural}, aims to find a useful pretrained model from a hub of various architectures that have been trained on different datasets.
Simpler methods do a wasteful two-stage optimization~\citep{pmlr-v139-you21b, nguyen2020leep}, or teat model selection as only a categorical hyperparameter~\citep{arango2024quicktune}, ignoring critical model details like architecture and pretrained weights.
DEHB~\cite{awad2021dehb} and ST-NAS~\cite{cai2021st} address these inefficiencies by integrating HPO and NAS processes but are not applicable for selecting from a pretrained hub.
In contrast, our model richly featurizes choices from a hub via their weights and avoids a two-stage process.

\vspace{-0.01\textheight}
\subsection{Learning Features from Weight Spaces}
\vspace{-0.005\textheight}

There has been a growing interest in techniques for processing neural network weights (and gradients).
A simple approach is flattening network parameters into a vector, which is effective for certain tasks \cite{unterthiner2020predicting}, but ignores inherent structures, such as symmetries within the parameter space.
For instance, permuting the neurons in the hidden layers of a multilayer perceptron does not alter the output \cite{hecht1990algebraic}.
Recent works that respect architectural symmetries have proven significantly more data-efficient \cite{peebles2022learning, lim2023graph, navon2023equivariant, zhou2023permutation}, particularly in tasks like predicting generalization from weights.
Leveraging these techniques, we featurize network weights for HPO by inputting a network's partially trained weights to a graph neural network, called a graph meta-network (GMN) for neural network inputs.
We do this because we seek a method for selecting models for finetuning, and the network weights encode information about the architecture, loss, training dataset, and optimization process.

\newpage
\vspace{-0.01\textheight}
\section{Limitations and Future Directions}
\vspace{-0.01\textheight}

FMS requires logged checkpoints, which can be expensive to store and communicate over networks or featurize with neural networks.
These weight checkpoints may only be available for a subset of hyperparameter evaluations, potentially limiting the effectiveness of our approach.
However, many HPO pipelines trivially log weights, and it should be feasible to make the weight input to our network optional, allowing flexibility in cases where checkpoint data is sparse.
We also introduce additional costs in training time, inference time, and memory usage for the PIGMN.

FMS was tested primarily on small- to medium-scale architectures within image classification tasks using small- to medium-sized datasets.
Further investigation is necessary to see if FMS generalizes to broader sets of unseen architectures, datasets, and tasks.
Our surrogate was trained on a limited-size HPO evaluation dataset due to compute and GP scalability constraints.
PIGMN weight features may be useful to scalable GP variants~\citep {wang2022scalable} or, more generally, learned surrogates~\citep{antoran2024scalable}.
We featurize fixed hyperparameter search spaces, and extending to dynamically changing spaces is an exciting avenue.

Other valuable information could be incorporated to enhance our model's performance, which current HPO methods neglect.
For instance, embeddings from large language models (LLMs) could process text related to the model's implementation, including code documentation, README files from GitHub repositories, and other relevant text data, which could provide context to improve HPO.

Our method, and all others in the BO framework, will struggle to tune more than tens to hundreds of hyperparameters, unlike gradient-based methods~\citep{lorraine2020optimizing}, which can tune millions of hyperparameters.
However, BO methods are more generally applicable in implementation, objective function choice, and hyperparameter search spaces.
BO methods struggle to tune some hyperparameter types -- such as when they are hierarchical or text-based.
FMS has similar complexities as other BO methods for selecting multiple configurations to evaluate in parallel~\citep{ginsbourger2010kriging, snoek2012practical}, unlike grid search.

We also use DyHPO's simplistic compute budget approximations, where the budget is known before evaluation and easily controllable, as it is simply the number of epochs.
This framework can be generalized to more realistic compute budget setups, such as when the cost of an HPO is not a parameter we directly control or when the cost is unknown before evaluation.

Allowing the HPO to choose from previously terminated optimization runs as in PBT~\citep{jaderberg2017population}, or, more generally, learning schedules of hyperparameters are routes to improved performance.
However, it is non-obvious how to integrate this into the DyHPO framework.

\vspace{-0.01\textheight}
\section{Conclusion}
\vspace{-0.01\textheight}

In this work, we propose Forecasting Model Search (FMS), a hyperparameter optimization method that uses checkpointed model weights to better guide our search.
We demonstrate that FMS performs well in selecting which model to train along with its fine-tuning hyperparameters.
By incorporating information from logged weight checkpoints, FMS provides another axis to enhance HPO with a large corpus of logged metadata from training runs across various datasets and architectures.
In the future, we envision leveraging these tools to create general HPO methods that efficiently tune a broad set of problems by wielding large amounts of -- often pre-existing -- optimization metadata.

\subsubsection*{Ethics Statement}
    The approach presented in this paper facilitates machine learning research and applications by making it easier to find performant hyperparameters.
    Designing more efficient HPO can help lower the cost of model training (e.g., time, compute, and environmental impact), making machine learning experiments easier for those in other disciplines.
    Overall, the benefits and risks are likely similar to those of other automated machine learning (AutoML) research.

\subsubsection*{Acknowledgements}
    NVIDIA's TAO Toolkit team also contributed to design choices, making FMS more practical and appealing.
    The Python community~\cite{van1995python, oliphant2007python} made the underlying tools, including PyTorch~\cite{paszke2017automatic}, PyTorch Geometric~\cite{fey2019fast}, GPyTorch~\cite{gardner2018gpytorch}, Matplotlib~\cite{hunter2007matplotlib}, and more.

\subsubsection*{Disclosure of Funding}
    NVIDIA funded this work.
    Jonathan Lorraine received funding from student scholarships at the University of Toronto and the Vector Institute, which do not directly support this work.
\nocite{*}
\bibliography{references}

\newpage
\appendix

\section{Appendix}

\vspace{-0.02\textheight}
\begin{table}[h]
\caption{Glossary and Notation}
\begin{center}
    \begin{tabular}{c c}
        \toprule
        \textbf{Term} & \textbf{Definition} \\
        \midrule
        HPO & Hyperparameter Optimization\\
        FMS & Forecasting Model Search\\
        GP & Gaussian Process\\
        EI & Expected Improvement\\
        BO & Bayesian Optimization\\
        DyHPO & Dynamic Multi-Fidelity Hyperparameter Optimization~\citep{wistuba2023supervising}\\
        CNN & Convolutional Neural Network\\
        GNN / GMN & Graph Neural Network / Graph Metanetwork\\
        PIGMN & Permutation-Invariant Graph Metanetwork~\citep{lim2023graph}\\
        ViT & Vision Transformer\\
        NFN & Neural Functional Network~\citep{zhou2023permutation}\\
        A100, H100 & NVIDIA A100/H100 Tensor Core GPU\\
        $i, k, d \in \mathbb{N}$ & Arbitrary indices\\
        $x, y, z \in \mathbb{R}$ &  Scalars\\
        $\mathbf{x}, \mathbf{y}, \mathbf{z} \in \mathbb{R}^{d}$ & Vectors\\
        $\mathbf{X}, \mathbf{Y}, \mathbf{Z} \in \mathbb{R}^{d \times d}$ & Matrices\\
        $\mathcal{X}, \mathcal{Y}, \mathcal{Z}$ & Sets\\
        $\mathbf{x} \in \mathcal{X}$ & A hyperparameter configuration and its domain \\
        $f:\mathcal{X} \to \mathbb{R}$ & Objective function which measures hyperparameter performance \\
        $y_i = f(\mathbf{x_i})$ & Observed performance for the $i^{th}$ hyperparameter \\
        $n \in \mathbb{N}$ & The total number of observed hyperparameter evaluations\\
        $\mathcal{D} = \{(\mathbf{x}_i, y_i)\}_{i=1}^n$ & Observed data \\
        $j \in \mathbb{N}$ & The compute budget\\
        $a$ & Acquisition function \\
        $\mu$ & Predictive mean \\
        $\sigma^2$ & Predictive variance \\
        $k$ & A kernel function \\
        $\mathbf{K}$ & Covariance matrix \\
        $\mathbf{\theta}$ & Kernel parameters \\
        $\mathbf{w}$ & Feature extractor network weights \\
        $\mathbf{Y}_{i,j-1}$ &  The learning curves for $i^{\textnormal{th}}$ evaluation and budget $j-1$\\
        $\psi$ & Feature extractor \\
        $\mathcal{L}$ & The loss function we optimize\\
        $\sigma$ & Activation function \\
        $\mathbf{W}$ & A checkpointed network weight\\
        $\ast$ & Graph convolution operation \\
        $L$ & The number of layers in the PIGMN\\
        $\Theta_k^{(l)}$ & PIGMN parameters at layer $l$ for kernel $k$ \\
        $\mathcal{G}$ & A graph \\
        $\mathcal{G}^0(\mathbf{W})$ & The graph for checkpointed weights $\mathbf{W}$ \\
        $\mathcal{G}^l$ & The graph at the $l^{\textnormal{th}}$ layer of the PIGMN \\
        $V(\mathcal{G})$ & The vertices or nodes of graph $\mathcal{G}$\\
        $v \in V$ & A specific vertex or node\\
        $\mathbf{h}_v^{l}$ & The feature vector of node $v \in V$ in at layer $l$\\
        $\xi$ & The features from the PIGMN\\
        \bottomrule
    \end{tabular}
\end{center}
\label{tab:TableOfNotation}
\end{table}

\subsection{Supplementary}

Our open-source code supports reproduction by providing all specific implementation details.
We have included scripts to generate the different hubs and scripts to run various experiments.
We describe several different variants of our methods when performing ablations of design choices, which our codebase supports.

\newpage
\subsection{Experimental Procedure}
\label{Experimental Procedure}

Our experiments used two model hubs, the \emph{Pretrained Model Hub} and \emph{Simple CNN Hub}.
The \emph{Simple CNN Hub} was taken from~\citet{unterthiner2020predicting}, and the procedure for generating \emph{Pretrained Model Hub} is detailed in Section~\ref{Generating the Pretrained Model Hub}.

After creating the model hubs, we make benchmarks of cached performance evaluations, used to simulate querying performance for computational efficiency while running our experiments, which is common for HPO experiments~\citep{arango2024quicktune, wistuba2023supervising}.
These benchmarks of cached performance evaluations were compiled for both \emph{Pretrained Model Hub} and \emph{Simple CNN Hub} that were fine-tuned on MNIST, CIFAR-10, and SVHN using hyperparameter settings specified in Section~\ref{Hyperparameter Settings}.

More details on the specifics of our acquisition function maximization can be found in Section~\ref{Acquisition Function Maximization}, and details of the HPO design choices (hyperhyperparameters) can be found in Section~\ref{meta-hyperparams}.

\subsubsection{Generating the Pretrained Model Hub}
\label{Generating the Pretrained Model Hub}

We created the \emph{Pretrained Model Hub}, a diverse collection of pretrained models for our experiments.
This hub features various architectures, including simple convolutional networks (CNNs), ResNets, and Vision Transformers (ViTs), each pretrained on ImageNet~\cite{imagenet}.

The pretraining procedure for ImageNet uses Stochastic Gradient Descent (SGD) with momentum.
The optimizer parameters are set with a momentum of $0.9$ and a weight decay of $1 \times 10^{-4}$.
The learning rate is initialized at $0.1$, with a step learning rate schedule, decaying the learning rate by a factor of $10$ every $30$ epochs.
The objective function is cross-entropy loss.
Training is performed for a total of $50$ epochs.
These settings are inherited from a standard PyTorch training example~\cite{pytorch2021imagenet}.

Using the hyperparameters outlined in Section~\ref{Hyperparameter Settings}, these architectures were trained on an NVIDIA A100 or an NVIDIA H100 GPU, depending on compute availability.
More details about the specific procedure are found in our open-source code.
Each model's pretraining took around $150$ minutes on an NVIDIA A100 and $75$ minutes on an NVIDIA H100 over $90$ epochs.
Additionally, each model fine-tuning took approximately $30$ minutes on an NVIDIA A100 and $15$ minutes on an NVIDIA H100 over $50$ epochs.
In total, we pretrained $4$ different main architectures, including ResNet~\citep{resnet}, ViT~\citep{vit}, CNN~\citep{726791}, and Deep Set~\citep{zaheer2017deep} and fine-tuned $50$ different hyperparameter configurations (of which model architecture is also a hyperparameter).
The resulting model hub provides a dataset for testing FMS's efficacy across various neural network architectures and initializations, enabling us to assess FMS's performance in selecting and tuning models from a heterogeneous set.

\subsection{Computational Considerations for Training Our Surrogate}
\label{Compute Requirements and Considerations}

FMS experiments were performed on an NVIDIA A100 GPU for $8$ hours for each HPO loop iteration (i.e., using the entire compute budget on a task).
DyHPO experiments took $30$ minutes per HPO loop iteration.
Our cost increases because the feature extractor includes a PIGMN, which we train during optimization.
During an HPO loop iteration, assuming a total compute budget of $100$ epochs, we often evaluate around $70$ different hyperparameter configurations.
Because the evaluation of configurations is cached in our benchmarks, the time per HPO loop iteration is primarily spent in training for the PIGMN.
The \emph{initial full training phase} lasts for $10$ hyperparameter evaluations to establish a strong initial set of surrogate model parameters, during which $1000$ optimization steps after each evaluation.
Later, in the \emph{refining phase}, we run $50$ optimization iterations after each evaluation.
The refinement phase lasts until the end of the HPO iteration loop and is meant to spend minimal effort to fine-tune the surrogate model parameters based on new data gathered.

\newpage
\subsection{Hyperparameter Search Space}
\label{Hyperparameter Settings}
The hyperparameters search spaces used for our experiments are listed below:

\begin{itemize}
    \item \textbf{Pretrained model index}: \([0, 30000]\) for \emph{Simple CNN Hub} and \([0, 4]\) for \emph{Pretrained Model Hub} since only four different architectures are pretrained.
    This parameter represents the architecture choice from a model hub.
    \item \textbf{Dropout}: \([0.0, 1.0]\).
    This parameter prevents overfitting by randomly setting a fraction of weights to \(0\) at each update during training.
    \item \textbf{Batch size}: \(\{16, 32, 64, 128, 256, 512\}\).
    This parameter determines the number of samples processed before the model is updated.
    \item \textbf{Learning rate}: \([1\text{e-4}, 1\text{e-1}]\), sampled on a logarithmic scale.
    Specifically, values are drawn uniformly in log-space and then exponentiated.
    \item \textbf{Momentum}: \(\{0.1, 0.5, 0.9\}\).
    The standard momentum parameter for SGD.
    \item \textbf{Weight decay}: \([1\text{e-5}, 1\text{e-1}]\), sampled on a logarithmic scale in the same manner as the learning rate.
\end{itemize}

These hyperparameters do not use explicit normalization or encoding.
This setup is inherited from DyHPO~\cite{wistuba2023supervising}.
Although we opted to run experiments with a limited search space of hyperparameters to reduce computation, our method can be extended to work with larger spaces of tens to hundreds of hyperparameters.

\subsection{Acquisition Function Maximization}\label{Acquisition Function Maximization}

We inherit the acquisition function maximization strategy of DyHPO and summarize it here for reference.
The acquisition function is maximized using a combination of random sampling and surrogate model predictions.
Initially, uniform sampling from the hyperparameter search space generates a set of candidate configurations.
The surrogate model predicts the expected performance and associated uncertainty for each candidate.
We use the predictions to compute the values of the multifidelity acquisition function as in Equation~\ref{eq:ei_acquistion}.

We evaluate each candidate's acquisition function and select the configuration with the highest value.
We then allocate an additional computational budget to the selected configuration.

\subsubsection{Budget Allocation}

We inherit the budget allocation strategy of DyHPO and summarize it here for reference.
We dynamically allocate resources based on intermediate performance evaluations to guide budget allocation.
Initially, all configurations are assigned a fixed budget of $1$ epoch, allowing a quick assessment of all configurations without consuming excessive resources.
Configurations that perform well receive incremental budget increases in subsequent evaluations by $1$ epoch each time, referred to as the \textit{fantasize step}.
When we evaluate a configuration a subsequent time, we resume from the checkpoint of its last configuration.
This strategy attempts to efficiently use computational resources by focusing on the most promising hyperparameter configurations.

\newpage
\subsection{Surrogate Function Design Choices}\label{meta-hyperparams}

We have various design choices for our surrogate model, which are sometimes called hyperhyperparameters or metaparameters, because they are parameters of the HPO.
Most design choices were inherited from DyHPO~\citep{wistuba2023supervising} and, for the PIGMN, from \citet{lim2023graph}.

For the surrogate, described in Figure~\ref{fig:fms-overview-tikz}, we use the following design choices:
We use the leaky ReLU activation function.
We use two hidden layers with $64$ and $128$ units, respectively.
We output $10$ features for the deep kernel GP.
The CNN architecture processing the learning curves uses two convolutional layers with $4$ and $8$ channels and a kernel size of $3$.
These features from our setup were inherited from DyHPO~\cite{wistuba2023supervising}.
Lastly, the PIGMN architecture comprises three layers with $64$, $128$, and $256$ units, respectively, which are defaults inherited from \citet{lim2023graph}.

The surrogate function was trained using Adam~\cite{kingma2014adam} with default parameters.
The model trains for $1000$ epochs in the initial training phase to fully learn from the initial points.
After, it switches to a refining phase, which trains for $50$ epochs per hyperparameter evaluation to fine-tune the model based on new data.
The number of refinement and initial full-training epochs was inherited from DyHPO~\cite{wistuba2023supervising}.
An epoch equals a gradient descent step since the entire dataset is used for each update.
Storing all checkpointed weights for each update in memory is impossible in a GPU, so we load checkpoints on the fly.
For each PIGMN update, we first compute and store the features for each hyperparameter evaluation.
Then, we approximate the inverse-matrix-vector product by solving a linear system using these cached features.
We do not explicitly compute the full kernel matrix $\mathbf{K}$ and its inverse, instead efficiently solving the linear system. 
Specifically, we use GPyTorch, which solves the linear systems internally during the training process~\cite{gardner2018gpytorch} using conjugate gradients.
This allows us to handle larger matrix computations for gradient updates without exceeding GPU memory limits.

\subsection{Additional Ablations}
\label{Additional Ablations}

\begin{figure}[ht!]
  \centering
  \begin{tikzpicture}
    \node (img) {\includegraphics[width=0.97\linewidth]{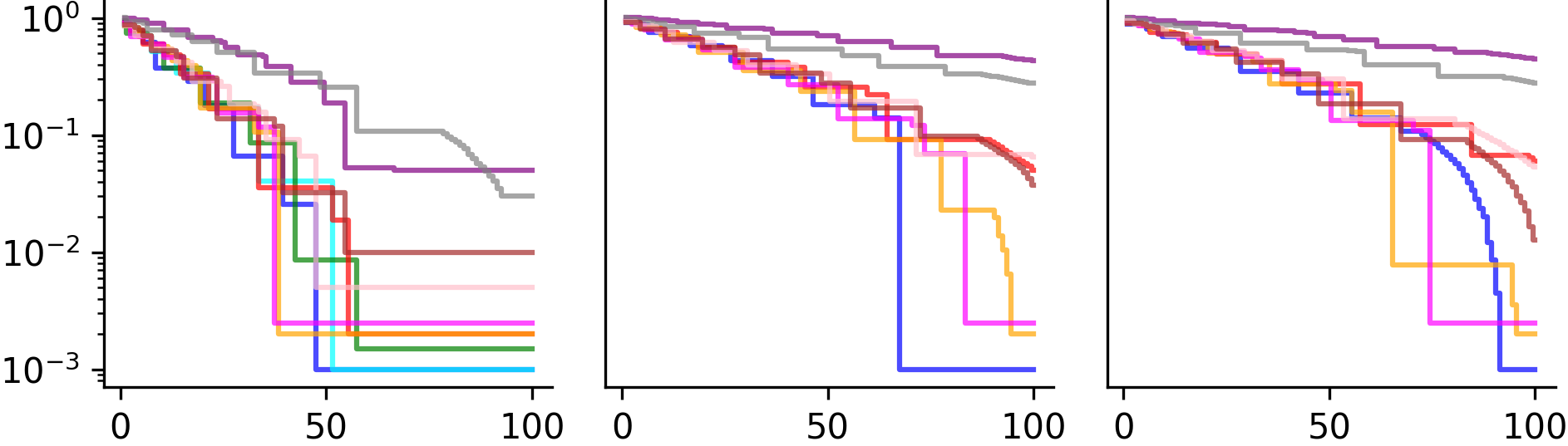}};
    \node[left=of img, node distance=0cm, rotate=90, xshift=0.65cm, yshift=-.9cm, font=\color{black}] {Regret};
    \node[below=of img, node distance=0cm, xshift=-.1cm, yshift=1.1cm,font=\color{black}] {Compute Budget in Epochs};
    \node[above=of img, node distance=0cm, xshift=-4.0cm, yshift=-1.0cm,font=\color{black}] {Simple CNN Hub};
    \node[above=of img, node distance=0cm, xshift=0.5cm, yshift=-1.0cm,font=\color{black}] {PTM Hub (SVHN)};
    \node[above=of img, node distance=0cm, xshift=5.0cm, yshift=-1.0cm,font=\color{black}] {PTM Hub (CIFAR-10)};
    
    \node[below=of img, node distance=2cm, yshift=0.4cm, xshift=0cm] {
      \begin{tikzpicture}
        \node (legend) [draw, fill=white, font=\small, inner sep=3pt] {
          \begin{tabular}{cccc}
            \textcolor{blue}{\rule{0.4cm}{0.2cm}} & FMS-GMN & \textcolor{cyan}{\rule{0.4cm}{0.2cm}} & FMS-NFN \\
            \textcolor{green}{\rule{0.4cm}{0.2cm}} & FMS-NFN (no CNN) & \textcolor{red}{\rule{0.4cm}{0.2cm}} & FMS-GMN (no CNN) \\
            \textcolor{orange}{\rule{0.4cm}{0.2cm}} & FMS-FLAT & \textcolor{magenta}{\rule{0.4cm}{0.2cm}} & FMS-FLAT (no CNN) \\
            \textcolor{darkorchid}{\rule{0.4cm}{0.2cm}} & Random Search & \textcolor{gray}{\rule{0.4cm}{0.2cm}} & DyHPO (no CNN) \\
            \textcolor{brown}{\rule{0.4cm}{0.2cm}} & DyHPO & \textcolor{pink}{\rule{0.4cm}{0.2cm}} & DyHPO with PTM index \\
          \end{tabular}
        };
      \end{tikzpicture}
    };
  \end{tikzpicture}
  \caption{
    We show the regret against the compute budget for the hyperparameter optimization (HPO) method across different hubs in each plot and various methods in each color.
    The regret values reflect the difference between the actual performance and the best possible performance over time.
    Lower regret indicates better performance.
    Our method, FMS-GMN, consistently shows lower regret over time across all hubs, demonstrating its effectiveness in HPO.
    The compute budget is measured in epochs (a full pass through the dataset), standardizing the compute effort across different tasks.
    FMS-NFN doesn't support diverse architectures, so it only runs on Simple CNN Hub.
  }
  \label{fig:regret_over_time_detailed}
\end{figure}
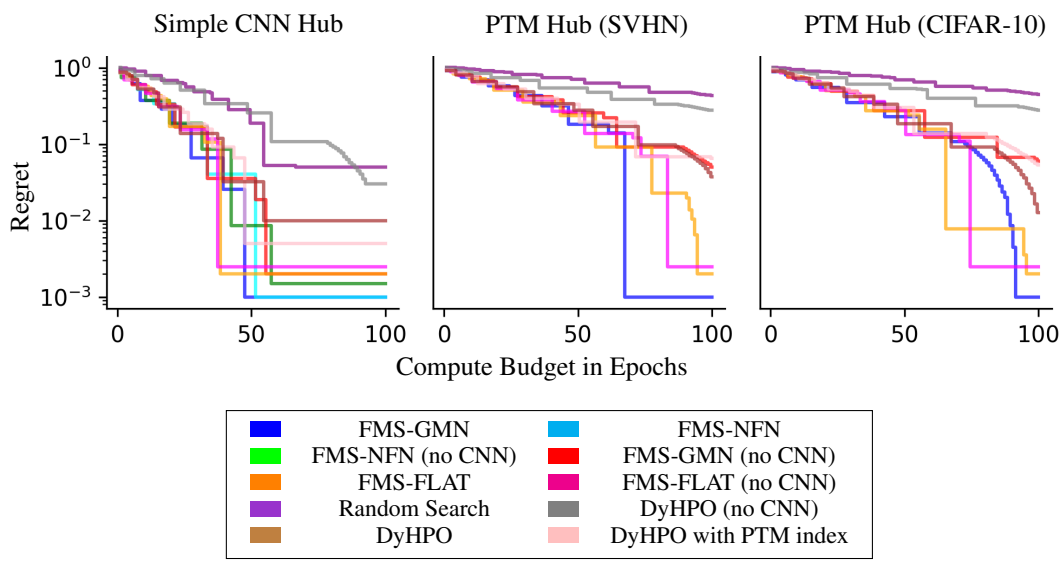

In Figure~\ref{fig:regret_over_time_detailed}, we perform ablations of FMS.
We study how permutation invariance affects FMS in FMS-FLAT, where we flatten the vector.
We also look at how incorporating features from the learning curve with a CNN interacts with different system parts.
Interestingly, while DyHPO without a CNN performs extremely poorly, FMS variants without a CNN still perform decently.
We speculate this is due to weight-space features that accommodate the lack of learning curve features.

\end{document}